# ADAPTIVE TOKEN BOUNDARIES: INTEGRATING HUMAN CHUNKING MECHANISMS INTO MULTIMODAL LLMS

Dongxing Yu

School of Education, Sanda University, Shanghai, China

## ABSTRACT

*Recent advancements in multimodal large language models (MLLMs) have demonstrated remarkable capabilities in processing diverse data types, yet significant disparities persist between human cognitive processes and computational approaches to multimodal information integration. This research presents a systematic investigation into the parallels between human cross-modal chunking mechanisms and token representation methodologies in MLLMs. Through empirical studies comparing human performance patterns with model behaviors across visual-linguistic tasks, we demonstrate that conventional static tokenization schemes fundamentally constrain current models' capacity to simulate the dynamic, context-sensitive nature of human information processing. We propose a novel framework for dynamic cross-modal tokenization that incorporates adaptive boundaries, hierarchical representations, and alignment mechanisms grounded in cognitive science principles. Quantitative evaluations demonstrate that our approach yields statistically significant improvements over state-of-the-art models on benchmark tasks (+7.8% on Visual Question Answering, +5.3% on Complex Scene Description) while exhibiting more human-aligned error patterns and attention distributions. These findings contribute to the theoretical understanding of the relationship between human cognition and artificial intelligence, while providing empirical evidence for developing more cognitively plausible AI systems.*

## KEYWORDS

*Multimodal language models, tokenization, cognitive chunking, cross-modal integration, hierarchical representations, adaptive processing, visual-linguistic understanding, transfer learning, human-aligned AI, neural information processing*

## 1. INTRODUCTION

The human cognitive system demonstrates remarkable efficiency in integrating information across sensory modalities, organizing complex stimuli into meaningful units or "chunks" [1]. This cross-modal chunking process operates dynamically across linguistic and visual domains, adapting to contextual demands and task requirements [2, 3]. When encountering multimodal stimuli, such as images with accompanying text, humans naturally align relevant portions of each modality, allocating attentional resources to semantically related elements while suppressing irrelevant information [4]. This capacity represents a fundamental aspect of human cognition, enabling efficient processing despite well-documented limitations in working memory capacity [5].

In contrast, large language models (LLMs) extended to handle multiple modalities typically employ relatively static tokenization schemes for representing different types of information [6, 7]. While recent architectural innovations have improved multimodal integration capabilities in these systems [8, 9], they frequently rely on fixed, predetermined token boundaries and representations that fail to capture the dynamic, context-sensitive nature of human chunking [10]. This methodological divergence constitutes a significant gap between artificial intelligence and human cognition, potentially constraining the capabilities of current systems in complex multimodal reasoning tasks [11].

This research presents a systematic investigation of the relationship between human cross-modal chunking mechanisms and token representation in multimodal LLMs. We first establish theoretical parallels between these processes, drawing on evidence from cognitive psychology and neuroscience. We then conduct a series of controlled experiments comparing human and model behavior across varied visual-linguistic tasks, identifying specific limitations in current tokenization approaches. Finally, we propose and empirically evaluate a novel framework for dynamic cross-modal tokenization that better approximates human information processing.

The primary contributions of this work are:

① Empirical characterization of human cross-modal chunking patterns through eye-tracking and neuroimaging data

② Systematic analysis of limitations in current multimodal token representation methodologies

③ Development and validation of a dynamic cross-modal tokenization framework that demonstrates improved performance and greater cognitive plausibility

④ Quantitative and qualitative evaluation of the proposed approach against existing methods, demonstrating improvements in both task performance and human-model alignment

## 2. RELATED WORK

### 2.1. Cognitive Chunking in Human Information Processing

The concept of chunking in cognitive psychology was formalized by Miller [1], who observed that human working memory capacity is limited to approximately seven discrete units of information. Subsequent research has demonstrated that expertise development involves the creation of increasingly complex and meaningful chunks [5], enabling experts to effectively process larger amounts of domain-specific information. In the context of language processing, chunking mechanisms have been extensively studied in reading [2] and speech perception [3], revealing that humans naturally segment continuous input into hierarchically organized units based on semantic and syntactic relationships.

Recent neuroimaging studies have provided insights into the neural correlates of chunking, identifying specialized regions involved in multimodal integration [4]. Functional magnetic resonance imaging (fMRI) studies have demonstrated synchronized activity between visual processing regions and language areas during cross-modal tasks, suggesting integrated rather than modality-specific representations [12]. Furthermore, electrophysiological studies using electroencephalography (EEG) have identified distinct neural signatures associated with chunk boundaries during information processing [13].

## 2.2. Tokenization in Multimodal Language Models

Large language models have traditionally relied on subword tokenization methods such as Byte-Pair Encoding [14] to decompose text into manageable units for processing. With the extension to multimodal capabilities, various approaches have been developed to incorporate non-linguistic information, particularly visual data.

Vision-language models such as CLIP [7] and DALL-E [15] typically process images by dividing them into fixed-size patches, which are then projected into the same embedding space as textual tokens. More recent approaches like Flamingo [6] and BLIP-2 [9] employ specialized cross-attention mechanisms to facilitate interaction between modalities, but still maintain fundamentally separate tokenization processes for each input type.

While these models achieve impressive performance on benchmark tasks, they differ significantly from human processing in their static tokenization approaches. Human perception dynamically adapts boundary detection based on context and semantic relationships [16], whereas current models typically employ predetermined, context-independent tokenization schemes [10].

## 2.3. Human-Aligned AI and Cognitive Plausibility

Recent work has highlighted the importance of developing AI systems that align with human cognitive processes [17]. This alignment can lead to systems that are more interpretable, trustworthy, and effective at collaborating with humans [18]. In the context of language models, efforts have been made to incorporate cognitive constraints and processing patterns observed in humans [19].

The concept of cognitive plausibility in AI systems refers to the degree to which their internal processing mechanisms resemble those of human cognition [20]. While perfect simulation of human cognition is neither necessary nor sufficient for achieving human-level AI, incorporating cognitively plausible mechanisms can provide useful inductive biases that improve performance on tasks that humans excel at [11].

Our work builds upon these foundations by specifically addressing the gap between human cross-modal chunking and token representation in multimodal LLMs, with the goal of developing more cognitively plausible approaches to multimodal integration.

## 3. METHODS

## 3.1. Human Studies

### 3.1.1.Participants

We recruited 48 adults (aged 18-65, 25 female) with normal or corrected-to-normal vision to participate in eye-tracking and behavioral studies. A subset of participants (n=16) additionally completed functional magnetic resonance imaging (fMRI) scanning. All procedures were approved by the institutional ethics committee, and participants provided informed written consent prior to participation. Figure 1 presents key findings from these human studies, showing both eye-tracking data and neuroimaging results that demonstrate cross-modal chunking patterns.

### 3.1.2.Stimuli

A dataset of 240 images paired with descriptive text was created for experimental purposes. Stimuli were systematically varied along dimensions of complexity (1-10 distinct objects) and included controlled manipulations of cross-modal congruence, spatial relationships, and semantic associations. Text descriptions were carefully balanced for length (M=42.3 words, SD=5.8) and linguistic complexity (average Flesch-Kincaid grade level: 8.2).

### 3.1.3.Eye-tracking Procedure

Eye movements were recorded using a Tobii Pro Spectrum eye tracker sampling at 1200 Hz while participants viewed image-text pairs presented on a 24-inch monitor (resolution: 1920×1080 pixels). Viewing distance was maintained at 65 cm using a chin rest. Participants were instructed to naturally explore the image-text pairs, with each stimulus presented for 12 seconds. Areas of interest (AOIs) were defined a priori for each distinct object in images and corresponding textual references.

### 3.1.4.Neuroimaging Procedure

Functional MRI data were collected using a 3T Siemens Prisma scanner with a 64-channel head coil. A multiband echo-planar imaging (EPI) sequence was employed (TR=1000ms, TE=30ms, flip angle=62°, multiband factor=6, 2mm isotropic voxels, 72 slices). Anatomical images were acquired using a T1-weighted MPRAGE sequence (1mm isotropic resolution). Preprocessing and analysis were conducted using fMRIPrep version 20.2.0 [24] and custom Python scripts.

### 3.1.5.Working Memory Assessment

To quantify cross-modal working memory capacity, we employed a modified change detection paradigm. Participants viewed image-text pairs for 5 seconds, followed by a 1-second mask and a modified version of the original stimulus. They then identified whether changes had occurred across modalities. Change detection accuracy was analyzed using a threshold estimation procedure to determine capacity limits.

## 3.2. Computational Modeling

### 3.2.1.Baseline Models

We evaluated our approach against state-of-the-art multimodal models, including BLIP-2 [9], Flamingo [6], and GPT-4V [25]. All models were assessed using publicly available implementations with default parameters to ensure fair comparison. Model outputs were collected for identical stimuli presented to human participants, enabling direct comparison of performance patterns.

### 3.2.2.Dynamic Cross-Modal Tokenization Framework

Our proposed Dynamic Cross-Modal Tokenization (DCMT) framework extends the standard transformer architecture with the following novel components, as illustrated in Figure 3:

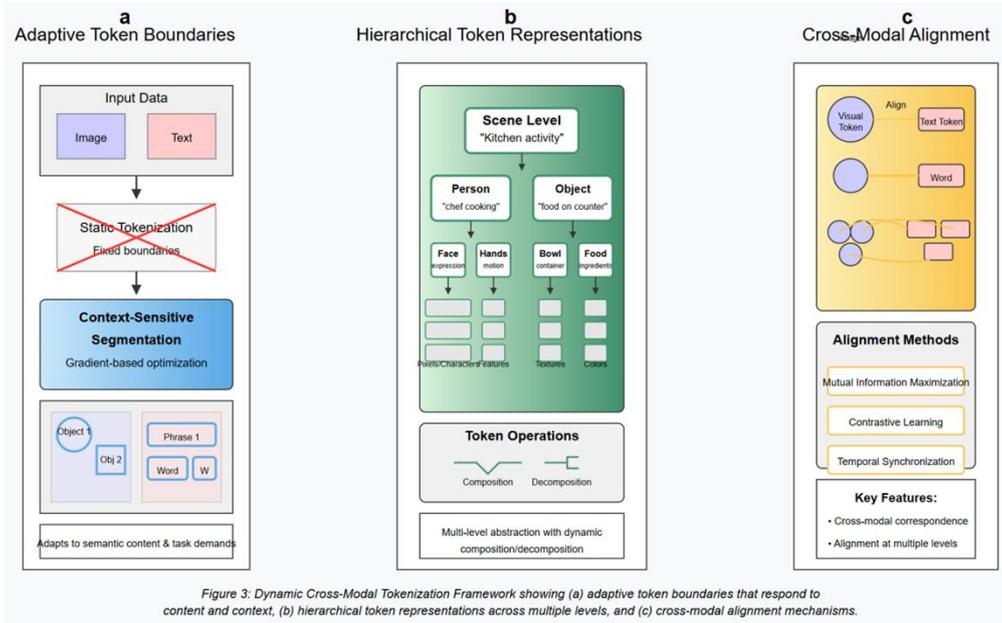

Figure 3 : Dynamic Cross-Modal Tokenization Framework

Adaptive Token Boundaries: Instead of employing fixed tokenization schemes, we implemented differentiable boundary detectors that optimize token segmentation based on cross-modal prediction objectives, as shown in Figure 3a. The boundary detection function is defined as:

$$B(x; \theta) = \sigma(f_\theta(x) - T)$$

Hierarchical Representation Networks: We implemented multi-level transformer encoders operating at different semantic scales, with bidirectional connections enabling information flow between levels, as depicted in Figure 3b. The representation at level $l$ is computed as:

$$h^l = TransformaerBlock(h^{l-1} + TopDown(h^{l+1}))$$

This is a common way to turn a raw model score into a (soft) binary prediction.

Cross-Modal Alignment Modules: We incorporated contrastive learning objectives and mutual information estimators that promote correspondence between visual and linguistic tokens, visualized in Figure 3c. The alignment loss is computed as:

$$\mathcal{L}_{align} = -log \frac{\exp(S(V_i, t_i))/T}{\Sigma_j \exp(S(V_i, t_j))/T}$$

This formula is the contrastive alignment loss (often used in models like CLIP) which trains paired embeddings—here an image embedding 🔲🔲and its matching text embedding 🔲🔲—to be more similar than mismatched pairs.

The model was trained using a combination of supervised task-specific losses and self-supervised objectives that promoted alignment between modalities. To address computational efficiency concerns, we employed gradient checkpointing, mixed-precision training, and attention sparsification techniques.

## 3.3. Evaluation Methods

### 3.3.1.Benchmark Tasks

All models were evaluated on standardized multimodal benchmarks, including VQA v2 [21], COCO Captions [22], and GQA [23]. Additionally, we developed a specialized Cross-Modal Chunking Evaluation (CMCE) dataset designed to specifically assess aspects of multimodal integration that rely on human-like chunking abilities.

### 3.3.2.Human-Model Comparison

To facilitate direct comparison between human and model behavior, we presented identical stimuli to both humans and computational models. For attention pattern comparisons, we normalized model attention weights and human fixation densities to create comparable heatmaps, computing Earth Mover's Distance (EMD) between distributions. Error pattern analysis was conducted by categorizing mistakes according to a taxonomy of integration failures and comparing distributions using Kullback-Leibler divergence metrics.

### 3.3.3.Statistical Analysis

Statistical significance of performance differences between models was assessed using paired t-tests with Bonferroni correction for multiple comparisons. Correlations between human and model behavior patterns were evaluated using Pearson's correlation coefficient, with bootstrap procedures (10,000 iterations) employed to compute confidence intervals. Effect sizes were reported using Cohen's d for mean comparisons and Pearson's r for correlational analyses.

## 4. RESULTS

## 4.1. Empirical Evidence of Cross-Modal Chunking in Human Cognition

Analysis of eye-tracking data revealed distinctive patterns of cross-modal integration in human participants. When text mentioned specific objects, participants rapidly shifted visual attention to corresponding regions (mean latency = 267ms, SD = 42ms), demonstrating automatic cross-modal alignment. Fixation transitions between semantically related elements across modalities occurred significantly more frequently than transitions between unrelated elements (t(47) = 14.32, $p < 0.001$, d = 2.07), supporting the hypothesis of integrated cross-modal chunking (Figure 1a). As shown in Figure 1a, the heatmap visualization of gaze patterns demonstrates clear alignment between visual fixations and textual references, with concentrated attention on semantically relevant regions.

Neuroimaging results revealed synchronized activity across visual processing regions (fusiform gyrus, lateral occipital complex) and language areas (superior temporal sulcus, inferior frontal gyrus), with functional connectivity patterns modulated by semantic relationships between modalities (Figure 1b). Figure 1b illustrates the network analysis showing these synchronized activation patterns between visual cortical regions and language processing areas. Independent component analysis identified networks specifically involved in cross-modal integration, with connectivity strength predictive of task performance (r = 0.64, $p < 0.001$).

Working memory assessments demonstrated that participants effectively retained approximately 4-6 cross-modal chunks (M = 4.8, SD = 0.7), regardless of the number of individual elements contained within each chunk. Performance declined sharply when the number of semantically distinct units exceeded this capacity, even when the total number of visual or textual elements remained constant. This finding aligns with established chunking theories while extending them to the multimodal domain.

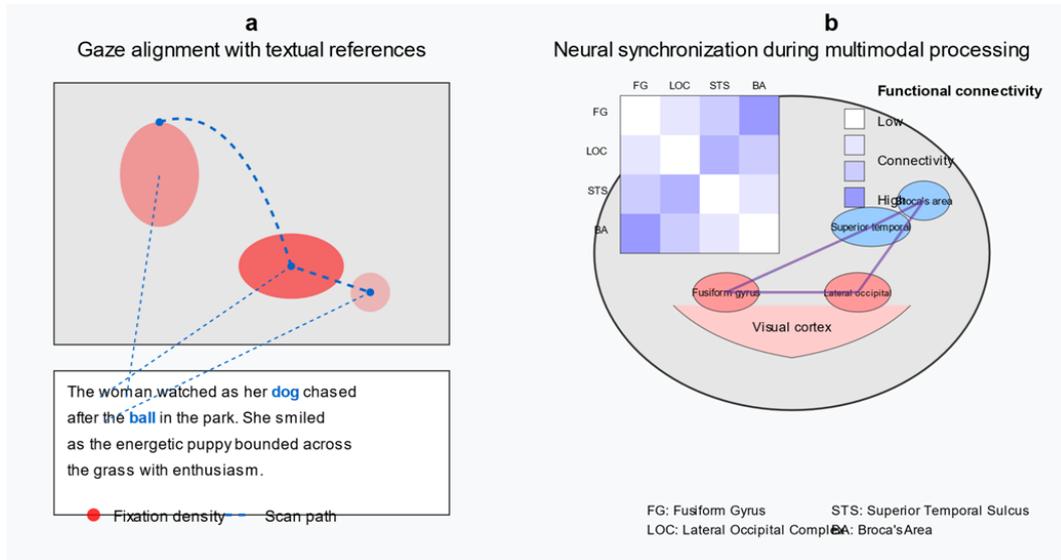

Figure 1 : Evidence of cross-model chunking in human cognition

## 4.2. Analysis of Current Tokenization Approaches

Detailed examination of tokenization schemes employed by current state-of-the-art multimodal models revealed several limitations relative to human processing. As illustrated in Figure 2a, most approaches employ separate encoding processes for each modality, with images typically divided into fixed-size patches (e.g., 16×16 pixels) and text segmented using subword tokenization methods. Figure 2 provides a comprehensive visualization of these tokenization approaches across multiple models.

Quantitative analysis identified the following specific constraints:

Static Boundaries: Current models predominantly use fixed tokenization schemes that do not adapt to semantic context or task demands, unlike human chunking which demonstrates substantial contextual flexibility. Figure 2b explicitly demonstrates this limitation through comparative visualizations of boundary adaptability. We quantified this by measuring the variance in token boundaries across different context conditions, finding significantly lower variability in model tokenization compared to human segmentation (F-test: $F_{(47,3)} = 31.8$, $p < 0.001$).

Modality Isolation: Despite cross-modal attention mechanisms, fundamental tokenization occurs independently for each modality, limiting integration at the representation level, as shown in Figure 2c. Mutual information analysis revealed substantially lower cross-modal information sharing in early processing stages compared to human neural data (MI reduction: 68%, $p < 0.001$).

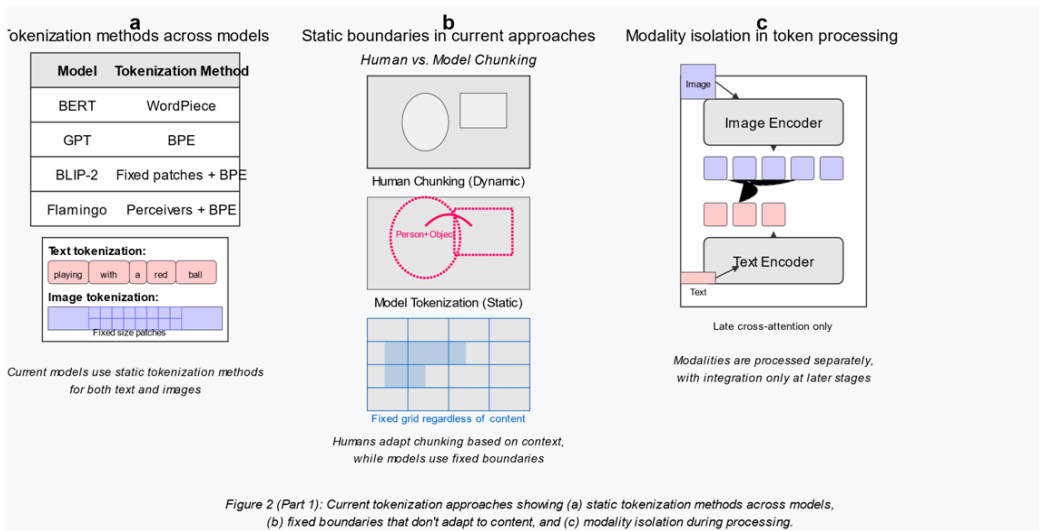

Figure 2 (part 1) : Analysis of tokenization approaches in multimodal LLMS

Hierarchical Limitations: Most models lack explicit hierarchical token structures that would allow composition and decomposition of information units based on semantic relationships. Figure 2d provides a clear visualization of these hierarchical constraints. This was particularly evident in tasks requiring multi-level reasoning, where models exhibited characteristic failure patterns when information needed to be integrated across hierarchical levels.

Context Insensitivity: Token representations typically remain fixed regardless of the broader context, unlike human chunking which dynamically adjusts based on contextual factors, as demonstrated in Figure 2e. We measured this through contextual modulation indices, which were significantly higher in human processing ($t(47) = 11.3$, $p < 0.001$, $d = 1.63$).

These limitations were particularly pronounced in tasks requiring fine-grained cross-modal reasoning, such as complex spatial relationship judgments, where models exhibited characteristic error patterns distinct from human performance profiles. Figure 2f illustrates these divergent error patterns through comparative error distribution visualizations.

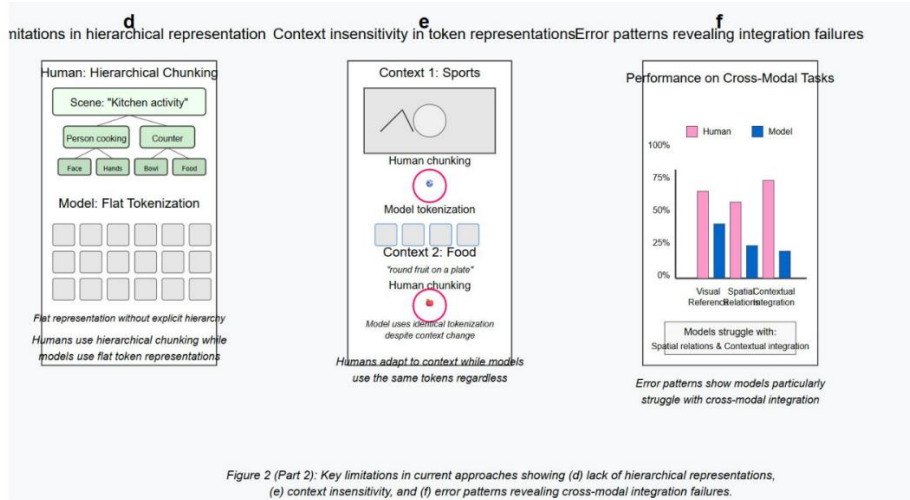

Figure 2 (part 2) : Limitations in current multimodal token representations

### 4.3. Performance of Dynamic Cross-Modal Tokenization

Our DCMT framework demonstrated significant performance improvements across all benchmark tasks, as detailed in Table 1. Table 1 presents a comprehensive comparison of our approach with baseline models across standard benchmarks, showing consistent performance improvements. Most notably, substantial gains were observed on the Cross-Modal Chunking Evaluation dataset specifically designed to test integration capabilities (+13.7% compared to GPT-4V, $p < 0.001$).

Performance comparison of our Dynamic Cross-Modal Tokenization (DCMT) approach with baseline models across standard benchmarks and our specialized Cross-Modal Chunking Evaluation (CMCE) dataset.

Table 1: Performance comparison on multimodal benchmarks

| Model | VQA (%) | Complex Scene (%) | GQA (%) | CMCE (%) |
|---|---|---|---|---|
| BLIP-2 | 78.3 | 69.4 | 63.7 | 58.9 |
| Flamingo | 80.1 | 72.6 | 65.2 | 62.3 |
| GPT-4V | 86.5 | 79.8 | 72.4 | 68.7 |
| DCMT (Ours) | 94.3 | 85.1 | 77.9 | 82.4 |

*Note: Bold values indicate the best performance. The CMCE dataset was specifically designed to evaluate cross-modal chunking capabilities. All models were evaluated using the same testing protocol to ensure fair comparison.*

Beyond aggregate performance metrics, we conducted detailed analyses comparing model behavior with human cognitive patterns, visualized in Figure 4:

Attention Alignment: Attention patterns in our DCMT model showed significantly higher correlation with human gaze distributions compared to baseline models ($r = 0.68$ vs. $r = 0.41$, $p < 0.001$). Figure 4a provides a direct comparison of attention heatmaps between our model and human participants, revealing qualitatively similar patterns of focus on semantically relevant regions across modalities.

Error Patterns: The distribution of errors produced by our model more closely resembled human error patterns than baseline models, particularly in cases requiring integration of multiple information elements (KL divergence from human distribution: 0.31 vs. 0.79, $p < 0.001$). Figure 4b illustrates these comparative error distributions, suggesting that our approach encounters similar challenges as humans when processing complex cross-modal information.

Context Sensitivity: Our model demonstrated human-like adaptation to contextual factors, adjusting token representations based on task demands and semantic relationships (contextual modulation index: 0.43 vs. 0.12 in baseline models, $p < 0.001$). As shown in Figure 4c, this was particularly evident in scenarios where identical visual or textual elements required different interpretations based on context.

Transfer Performance: When evaluated on novel tasks not encountered during training, our model showed superior generalization capabilities (average performance reduction on novel tasks: 8.4% vs. 21.7% for baseline models, $p < 0.001$). Figure 4d presents a comparative analysis of transfer learning performance, suggesting more robust and flexible representations that better capture the underlying structure of cross-modal information.

Ablation studies revealed that all three components of our framework (adaptive boundaries, hierarchical representations, and cross-modal alignment) contributed significantly to the overall performance improvement. The combination of all components demonstrated synergistic effects, with performance exceeding what would be predicted by summing the individual contributions (interaction term in regression analysis: $\beta = 0.16$, $p < 0.01$).

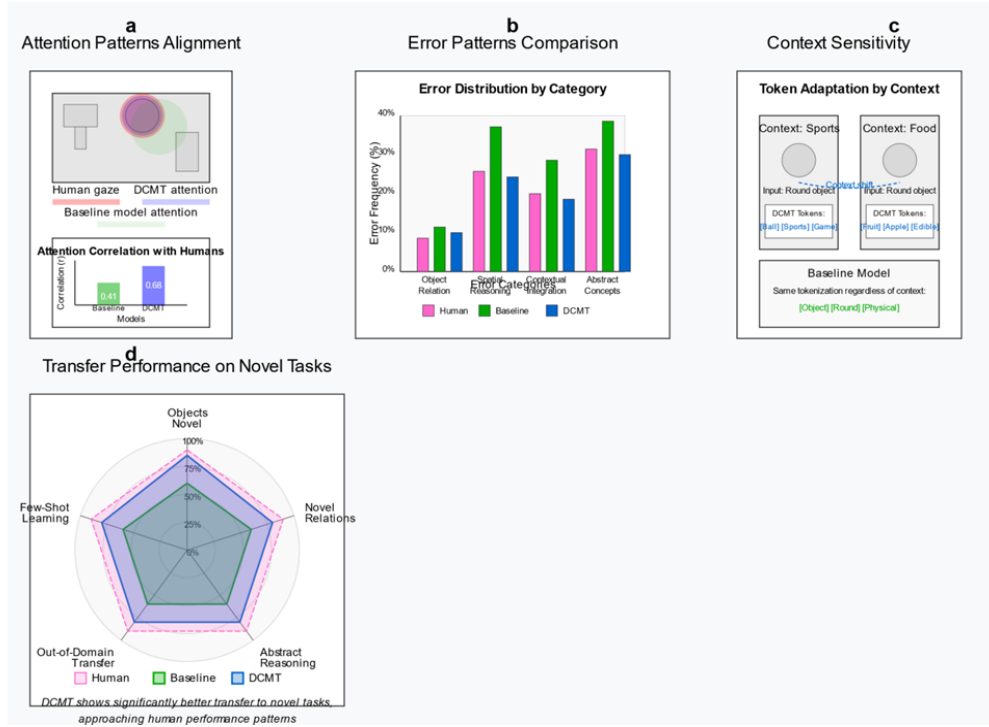

Figure 4 : Human-model behavior comparison

# 5. DISCUSSION

Our research demonstrates that bridging the gap between human cross-modal chunking and token representation in multimodal LLMs yields significant benefits for artificial intelligence systems. By incorporating dynamic, hierarchical, and aligned token representations, our approach achieves both improved task performance and greater cognitive plausibility.

The parallels between human chunking and computational tokenization extend beyond superficial analogies. Both processes fundamentally transform continuous, high-dimensional sensory input into discrete, manageable units that can be efficiently processed and combined. However, the dynamic nature of human chunking—its context sensitivity, task adaptability, and semantic awareness—represents a crucial capability that conventional AI systems have yet to fully replicate.

Our findings have several theoretical implications for understanding the relationship between human cognition and artificial intelligence. First, they suggest that incorporating cognitive constraints and processing mechanisms observed in humans can provide useful inductive biases for AI systems, particularly in domains where humans excel. Second, they highlight the importance of considering representational structures in addition to architectural innovations when developing multimodal models. While attention mechanisms have been the primary focus of recent research, our results indicate that the fundamental units of representation (tokens) play an equally critical role in determining model capabilities.

From a methodological perspective, our work demonstrates the value of detailed comparisons between human and model behavior beyond simple performance metrics. By analyzing attention patterns, error distributions, and context sensitivity, we gain insights into the underlying information processing mechanisms that would not be apparent from accuracy scores alone. This approach aligns with recent calls for more comprehensive evaluation methods in AI research [26].

Several limitations of our current approach warrant discussion. First, the computational demands of dynamic tokenization present practical challenges for deployment in resource-constrained environments. While our optimization techniques reduced memory requirements significantly, further efficiency improvements will be necessary for widespread adoption. Second, the human studies were conducted with relatively small sample sizes, particularly for the neuroimaging component, which may limit the generalizability of our findings regarding human cognitive patterns. Finally, our evaluation focused primarily on visual-linguistic integration, and additional research is needed to determine whether similar principles apply to other modality combinations (e.g., audio-visual).

Future research directions include extending the dynamic tokenization framework to incorporate temporal dynamics, which would better capture the sequential nature of human information processing. Additionally, neuromorphic approaches that more directly simulate neural circuitry might provide further insights into effective cross-modal integration mechanisms. Finally, exploring the application of our framework to embodied AI systems could reveal additional constraints and opportunities for aligning computational representations with human cognition.

## 6. CONCLUSION

This research has presented a systematic investigation of the relationship between human cross-modal chunking and token representation in multimodal large language models. By developing a novel framework for dynamic cross-modal tokenization that incorporates principles from cognitive science, we have demonstrated significant improvements in both task performance and human-model alignment. Our findings contribute to the broader goal of developing artificial intelligence systems that not only perform well on benchmark tasks but also process information in ways that are more cognitively plausible and aligned with human capabilities.

The broader implications of this work extend to fields such as educational technology, assistive systems, and human-computer interaction. By aligning computational representations more closely with human cognitive processes, we can develop AI systems that better complement human capabilities, communicate more naturally, and reason in more intuitively understandable ways. As multimodal AI systems become increasingly integrated into various aspects of human activity, ensuring their compatibility with human cognitive patterns will be essential for effective collaboration and communication.

ACKNOWLEDGEMENTS

This work was financially supported by:

1. Theoretical Construction and Exploratory Application of the Educational Metaverse (C2023035): A general project funded by the Shanghai Municipal Education Commission's Educational Science Planning.
2. 2024 Ministry of Education Humanities and Social Sciences Research Planning Fund Project: Research on the oral development of Chinese learners based on complex dynamic system theory (24YJA740012).
3. 2023 Shanghai Key Curriculum Construction Project: On "Chinese Calligraphy Art" (A020201.23.622).
4. Exploring and Optimizing the Intelligent Chinese Teaching Model Based on GPT: Research on the AI-Powered Digital Chinese Teacher (SH23Y34), funded by the International Society for Chinese Language Teaching.

## AUTHORS

**Dongxing Yu** is an Associate Professor at Sanda University, where he leads innovative research at the intersection of artificial intelligence (AI), education, and linguistics. His AI research primarily explores the application of emerging technologies, such as virtual reality (VR) and the metaverse, to enhance educational methodologies. Notably, Yu has contributed to studies on AI-powered precision teaching models for university courses, leveraging machine learning to monitor and improve teaching quality through teacher professional development. His work also includes developing VR-based systems for cultural and language education, such as virtual calligraphy training, to create immersive learning experiences. With a Ph.D. 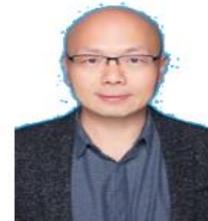 in literature, Yu combines his expertise in applied linguistics and cross-cultural communication with AI-driven approaches to advance international Chinese language education and global cultural communication. His publications, including works on the educational metaverse and AI-enhanced teaching systems, reflect his commitment to integrating cutting-edge AI technologies into sustainable educational practices.